\documentclass[conference]{IEEEtran}
\IEEEoverridecommandlockouts
\usepackage{cite}
\usepackage{boldline} 
\usepackage{amsmath,amssymb,amsfonts}
\usepackage{algorithmic}
\usepackage{graphicx}
\usepackage{textcomp}
\usepackage{xcolor}
\def\BibTeX{{\rm B\kern-.05em{\sc i\kern-.025em b}\kern-.08em
    T\kern-.1667em\lower.7ex\hbox{E}\kern-.125emX}}
\begin{document}

\title{Generating Contextual Load Profiles Using\\
 a Conditional Variational Autoencoder\\

\thanks{This work is supported by the Chinese Scholarship Council.}
}

\author{\IEEEauthorblockN{Chenguang~Wang, Simon~H.~Tindemans, Peter~Palensky\\}
\textit{Department of Electrical Sustainable Energy}\\
\textit{Delft University of Technology}\\
Delft, The Netherlands \\
\{c.wang-8,  s.h.tindemans, p.palensky\}@tudelft.nl}


\IEEEoverridecommandlockouts

\IEEEpubid{\parbox{\columnwidth}{\copyright 2022 IEEE. Personal use of this material is permitted. Permission from IEEE must be obtained for all other uses, in any current or future media, including reprinting/republishing this material for advertising or promotional purposes, creating new collective works, for resale or redistribution to servers or lists, or reuse of any copyrighted component of this work in other works.}\hspace{\columnsep}\makebox[\columnwidth]{ }}


\maketitle

\IEEEpubidadjcol

\begin{abstract}

Generating power system states that have similar distribution and dependency to the historical ones is essential for the tasks of system planning and security assessment, especially when the historical data is insufficient. In this paper, we described a generative model for load profiles of industrial and commercial customers, based on the conditional variational autoencoder (CVAE) neural network architecture, which is challenging due to the highly variable nature of such profiles.
Generated contextual load profiles were conditioned on the month of the year and typical power exchange with the grid. Moreover, the quality of generations was both visually and statistically evaluated. The experimental results demonstrate our proposed CVAE model can capture temporal features of historical load profiles and  generate `realistic' data with satisfying univariate distributions and multivariate dependencies.
\end{abstract}


\section{Introduction}

For power system planning and security assessment, it is of great significance to examine system performance through abundant scenarios \cite{bloomfield2021quantifying, panciatici2012operating}. However, when historical data is scarce or larger data sets are required for more precise analysis, it is crucial to building generative models for reproducing unlimited non-repeating data with similar marginal distributions and multivariate dependencies to historical data.

Parametric methods, such as hidden Markov models \cite{zia2011hidden} and \emph{Gaussian mixture models} (GMM), \cite{kang2007load} have been utilized to describe historical data patterns. Recently, vine-based copula models have been proposed (e.g., in \cite{Konstantelos2019}) to capture marginal distributions and high-dimensional dependencies of historical power system states. However, vine-based copula models are naturally asymmetric and have hard-to-quantify training bias due to sequential model selection.  

With the development of machine learning technologies, data-driven generative models, such as \emph{variational autoencoder} (VAE) \cite{kingma2013auto}, have been proposed to learn features of high-dimensional historical data and then create `unseen' ones. On this basis, conditional VAE (CVAE) \cite{sohn2015learning} made it possible to generate data under specific conditions. In \cite{wang2021generating}, the impact of the CVAE model's output noise on its generative performance has been investigated with a use case of learning and generating snapshots of country-level load states \cite{Muehlenpfordt2019}. However, such snapshots of large load aggregations have limited diversity and variability. 

In this paper, we bridge the gap by investigating the CVAE model's capacity to generate synthetic load profiles that are representative of those from a large variety of individual users. Compared to \cite{wang2021generating}, this work aims to generate consumption patterns that are both temporal (instead of spatial) and at a lower aggregation level, where the loads are more stochastic. The main contributions of this paper are as follows:

\begin{itemize}

\item We analyze the properties of daily load profiles of an anonymized data set of 5,000 industrial and commercial customers.

\item For better training and generation performance, we introduce data split,  month condition, and power exchange intensity calculation strategies during data processing.

\item We evaluate the performance of the CVAE model under different time (month) and power exchange intensity conditions with both visual and statistical metrics.

\end{itemize}

\section{Data Generation Mechanism}{\label{sec:Mechanism}}

In this section, a representative multivariate load state generation mechanism was described, based on the \emph{conditional variational autoencoder} (CVAE). The description summarized that in \cite{wang2021generating}.

\subsection{CVAE-based generative model}

The CVAE is a neural network architecture that is trained to learn the salient features of historical data by mapping (\emph{encoding}) historical system states onto a lower-dimensional latent space where the latent distribution is approximately normal - and transforming latent vectors back (\emph{decoding}) into a high-dimensional state space \cite{Kingma2019}. The decoder is used in conjunction with contextual information $c$ to generate representative states (which can be omitted to obtain a regular VAE model). Consequently, the model is able to generate samples with a similar distribution to the historical data by transforming normally distributed samples in the latent space back to the data space. We note that the latent (i.e., hidden) representation of a data point is used solely to facilitate reconstruction and synthesis. It does not need to be imbued with a particular meaning.

\begin{figure*} 
  \centering 
    \includegraphics[scale=0.56]{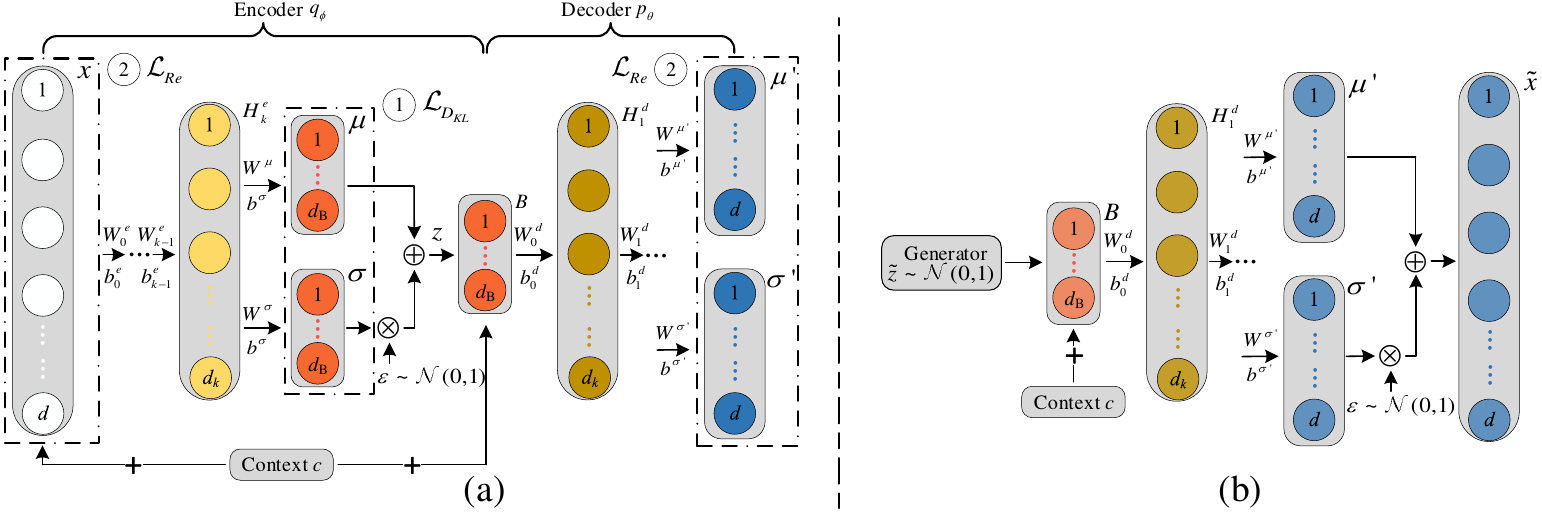}
  \caption{Schematic of a CVAE model. (a) The structure of a CVAE model for training. (b) The structure of a CVAE model when it's utilized as a generator.}
  \label{fig:CVAE}  
\end{figure*}

In the training process, the specific structure of the CVAE algorithm is depicted in Fig.~\ref{fig:CVAE}a. The \emph{encoder} maps the $d$-dimensional input data $x$ to the code $z$ in the lower-dimensional latent space through $k$ hidden layers $H_l^e$, $l=1,\ldots,k$. Weight matrices $W_{l}^{e}$, bias vectors $b_{l}^{e}$ and the context $c$ are utilized in the encoding process as
\begin{subequations}
		\begin{align}\label{eq:y}
		\left(\begin{matrix} \mu \\ \sigma \end{matrix}\right)  = & \left(\begin{matrix} 	
		W^{\mu} \\ W^{\sigma} \end{matrix}\right)(a(W_{k}^e(\ldots 	
		a(W_{1}^{e}(x,c) + b_{1}^{e})\ldots)+b_{k}^e)) \nonumber \\ &
		+ \left(\begin{matrix} b^{\mu} 	\\ b^{\sigma} \end{matrix}\right)\,,
		\\
	\label{eq:hatz}
	z =& \mu+ \epsilon \odot \sigma \, , 
	\end{align}
\end{subequations}
where $a$ represents an element-wise nonlinear activation function. Vectors $\mu$ and $\sigma$ parameterize an input-dependent normal distribution in the latent space. The output $z$ is sampled accordingly, using $\epsilon$, a vector that is sampled from a standard normal distribution, and the Hadamard product $\odot$. 
Mirroring the encoder network, the decoder maps the sampled latent space code $z$ to the $d$-dimensional data $\mu'$ and $\sigma'$ using
\begin{align}\label{eq:mu_2}
	\left(\begin{matrix} \mu' \\ \sigma' \end{matrix}\right) = & \left(\begin{matrix} W^{\mu'} \\ W^{\sigma'} \end{matrix}\right)(\ldots a(W_{1}^{d}(z,c) + b_{1}^{d})\ldots) + \left(\begin{matrix} b^{\mu'} \\ b^{\sigma'} \end{matrix}\right)\, ,
\end{align}
where $W_{l}^{d}$ and $b_{l}^{d}$ denote weight matrices and bias vectors for decoding, respectively. $\mu'$ and $\sigma'$ parameterize a $z$-dependent normal distribution in the $x$ space. 

After the training process, only the decoder part of the trained CVAE network is utilized to generate data. Latent space codes $\Tilde{z}$ are sampled from the standard normal distribution $\mathcal{N}(0,I)$ (see Fig.~\ref{fig:CVAE}b). Then, data space samples $\Tilde{x}$ are sampled from distribution $\mathcal{N}(\mu'(\Tilde{z},c),\sigma'(\Tilde{z},c))$ as $\tilde{x} = \mu' + \epsilon \odot \sigma'$, whose parameters are determined by $\Tilde{z}$ and $c$. 

\subsection{Optimization goal}
During training, weight matrices $W$ and bias vectors $b$ are updated iteratively to minimize the loss function \cite{Kingma2019}
\begin{align}\label{eq:loss_sum}
    \mathcal{L}=\mathcal{L}_{D_{KL}}+\mathcal{L}_{Re}.
\end{align}
The \emph{Kullback-Leibler loss} $\mathcal{L}_{D_{KL}}=\sum_i D_{KL}(q_\phi(z|x_i)||p(z))$ is the sum over all training data points $x_i$ (assumed i.i.d.) of the Kullback–Leibler divergence between that point's posterior distribution $q_\phi(z|x_i)$ and the prior distribution $p(z)$ (chosen as the standard normal distribution). The \emph{reconstruction loss} $\mathcal{L}_{Re}$, representing the negative log-likelihood of reconstructing the inputs $x_i$ via their latent space codes and the decoder that is parameterized by $\theta$, is written as $-\sum_{i=1}^{n} \mathbb{E}_{Z\sim q_\phi(z|x_i)}[\log_{P_\theta}(x_i|Z)]$. With a constant $\frac{n d}{2}\log2\pi$ omitted, the $\mathcal{L}_{Re}$ is computed as
\begin{align}\label{eq:Re_loss}
    \mathcal{L}_{Re}  \approx \frac{1}{2}\sum_{i=1}^n\sum_{j=1}^d((x_{i,j}-\mu'_{i,j})^2/\sigma'^2_{i,j}+\log\sigma'^2_{i,j}),
\end{align} 
where $n$ denotes the total number of observations used for training. 
During training, the full-sample sum in loss functions $\mathcal{L}_{D_{KL}}$ and $\mathcal{L}_{Re}$ are replaced by sample batch averages.
Inspired by \cite{wang2021generating}, the sample-dependent output noise parameter $\sigma'$ is co-optimised during training, and the noise $\epsilon \odot \sigma'(\Tilde{z},c)$ is used in the generative process. Also, a weighing factor $\beta$ was multiplied with $\mathcal{L}_{D_{KL}}$ to adjust the ratio between two losses in \eqref{eq:loss_sum} as $\mathcal{L}=\beta\mathcal{L}_{D_{KL}}+\mathcal{L}_{Re}$ \cite{burgess2018understanding}.

\section{Study description}
We used the CVAE-based generative model described above to generate daily load profiles (24 hours) of individual network connections (i.e., users), conditioned on the month of the year and power the user typically exchanges with the grid. The performance of the model was analyzed using a load data set of 5,000 users. The quality of generations was evaluated visually as a function of conditioning parameters. In addition, performance was validated statistically by measuring univariate distributions and multivariate dependencies. Moreover, an experiment was conducted to test the model capacity of interpolation.

\subsection{Data source}
\begin{figure}
  \centering 
    \includegraphics[scale=0.32]{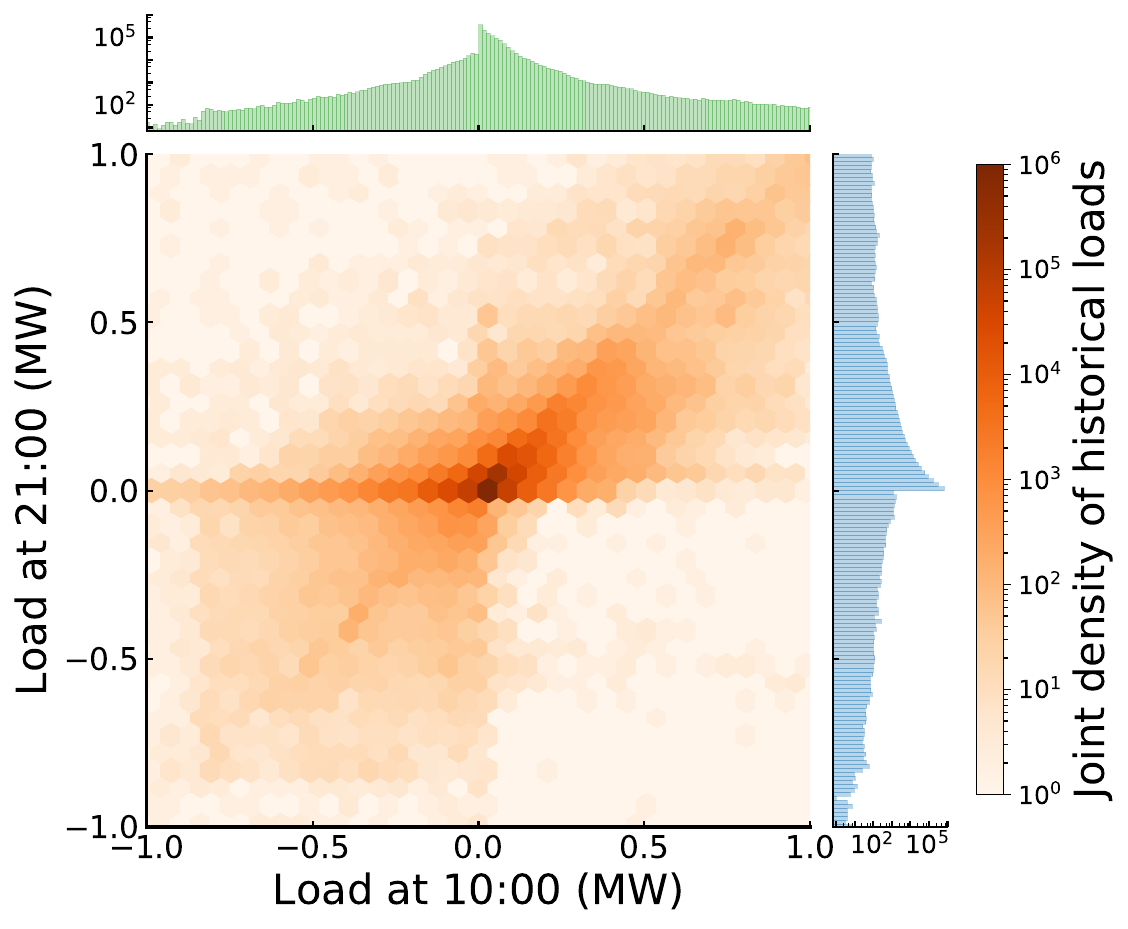}
  \caption{Marginal histogram and joint density of historical loads at 10:00 and 21:00 during one year.}
  \label{fig:Density-Pdf}  
\end{figure}

Anonymized historical electricity consumption/generation data of 5,000 industrial and commercial electricity users during 2020 was obtained from Alliander NV\cite{Alliander2022}, a Dutch distribution network owner and operator. The time resolution of the data is 15 minutes. It is worth noting that the data set's time label is \emph{UTC} (Coordinated Universal Time). However, the actual local time for electricity users is \emph{CET} (Central European Time). During standard time and daylight saving time, their time differences are 1 and 2 hours, respectively. The energy data was converted from integer kWh values to average power with multiples of 4~kW. 
Compared with country-level load profiles\cite{Muehlenpfordt2019}, the energy consumption of individual users involves more variability and less predictability. Fig.~\ref{fig:Density-Pdf} illustrates the large variety of daily profiles, by plotting the marginal histogram and joint density of all historical load profiles at 10:00 and 21:00. Note the logarithmic density used, indicating a large concentration around (relatively) small values. Moreover, data points located in the upper-left and bottom-right corners stand for users can not only consume but also generate energy. All these factors above make it challenging for the CVAE model to capture the load patterns.

\subsection{Data process}
\begin{figure}
  \centering 
    \includegraphics[scale=0.79]{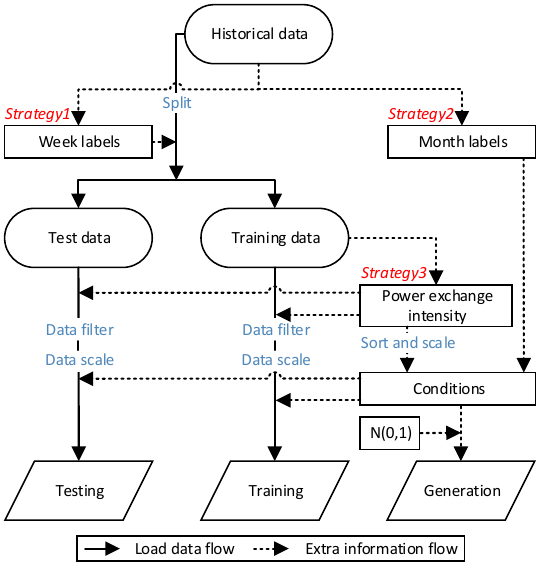}
  \caption{Data processing scheme.}
  \label{fig:Flow Chat}  
\end{figure}

The data processing scheme is shown in Fig.~\ref{fig:Flow Chat}. The historical data were split, scaled, and conditioned. Three data process strategies used in this study are as follows.

\subsubsection{Strategy I - Data split}
The historical data were randomly split into training and test sets as blocks of one week with a proportion of 4:1. This strikes a balance between separating individual days (subsequent days are not sufficiently independent) and separating larger blocks (insufficient coverage in the test set).

\subsubsection{Strategy II - Month Conditions}
In this study, one of the conditions (contextual information) $c$ is the month of the year. We used the $\sin{(\cdot)}$ and $\cos{(\cdot)}$ values of a month as the condition of load data. For a specific month $m$, its condition was encoded as $\sin{(\frac{m}{12}\cdot{2\pi})}$ and $\cos{(\frac{m}{12}\cdot{2\pi})}$. This encoding reflects the continuity and circularity of this feature. 

\subsubsection{Strategy III - User intensity}
After inspecting historical load profiles, we noticed that some users had relatively regular load profiles, whereas others had irregular behavior with rare consumption or generation spikes. Some connections were only active during a small part of the year. To construct a conditioning feature that represents the `size' of electricity users, we calculated the daily average power exchange by averaging over all non-zero values of the  \emph{absolute} power (consumption or generation) for each user and each day. For each customer, this value was averaged over the five days with the largest daily average power exchange to obtain the \emph{user intensity}. The intensity values were used to assign to each customer as a rank order $c \in [0,1]$. 
Due to the large range of power values present in the data (see Fig.~\ref{fig:Density-Pdf}) and the relative scarcity of data with high peak exchange, we trained the model only on profiles of customers with an intensity up to 100~kW. Ultimately, 4,049 users remained, with 1,170,110 and 307,720 load profiles in the training and test sets, respectively. The values were scaled by $1/(100~kW)$ for training. 

\subsection{Training and Data Generation}

\begin{figure}
  \centering 
    \includegraphics[scale=0.24]{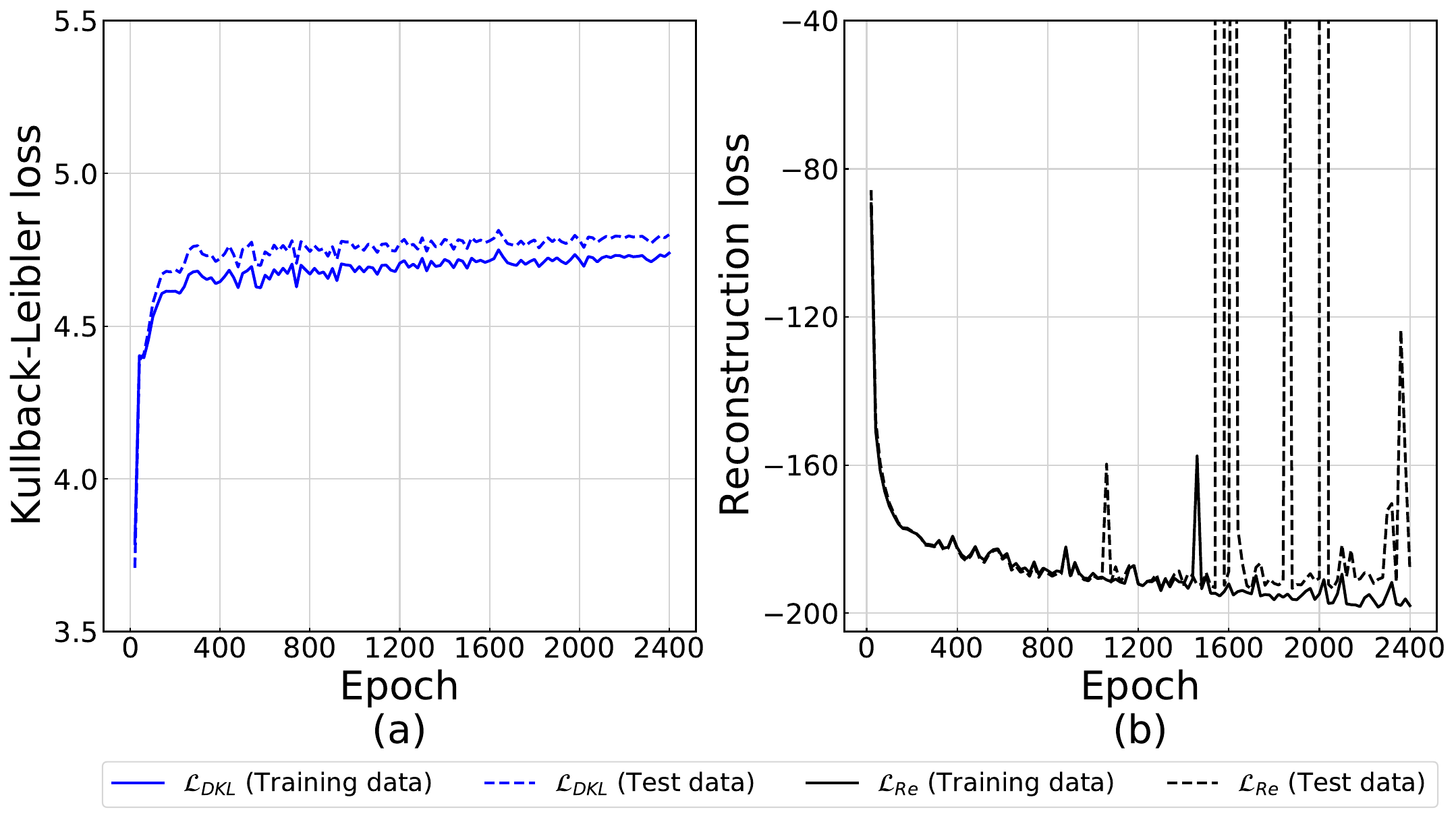}
  \caption{Training process and its failure. (a) Training process of Kullback-Leibler loss. (b) Training process of reconstruction loss.}
  \label{fig:ISGT-training_tendency}  
\end{figure}

\begin{figure*}
    \centering
	\includegraphics[scale=0.20]{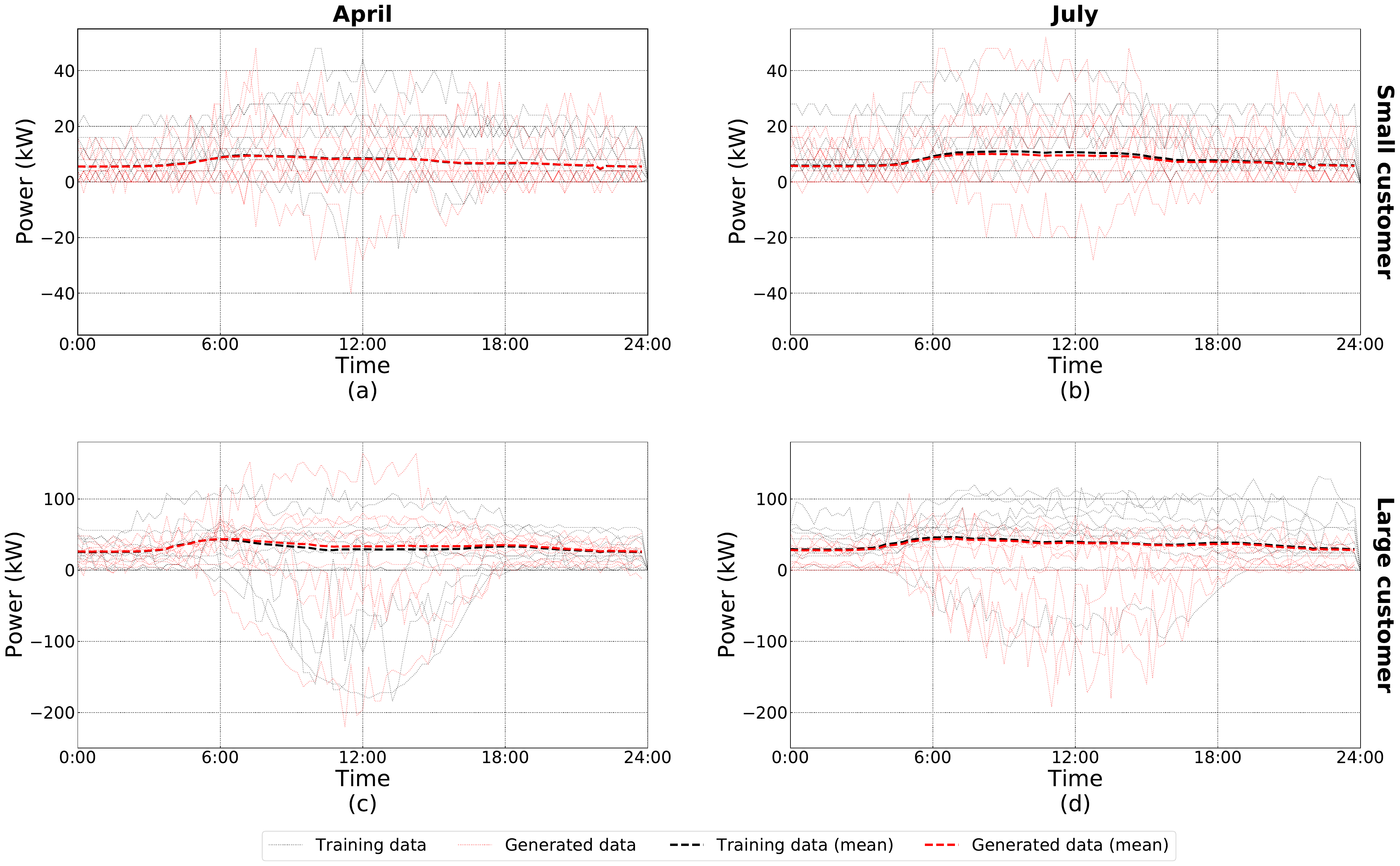}
	\caption{10 randomly sampled historical and generated load profiles and average daily load from customers of various sizes in different months.}
	\label{fig:Load profiles}  
\end{figure*}

\begin{figure*}
    \centering
	\includegraphics[scale=0.155]{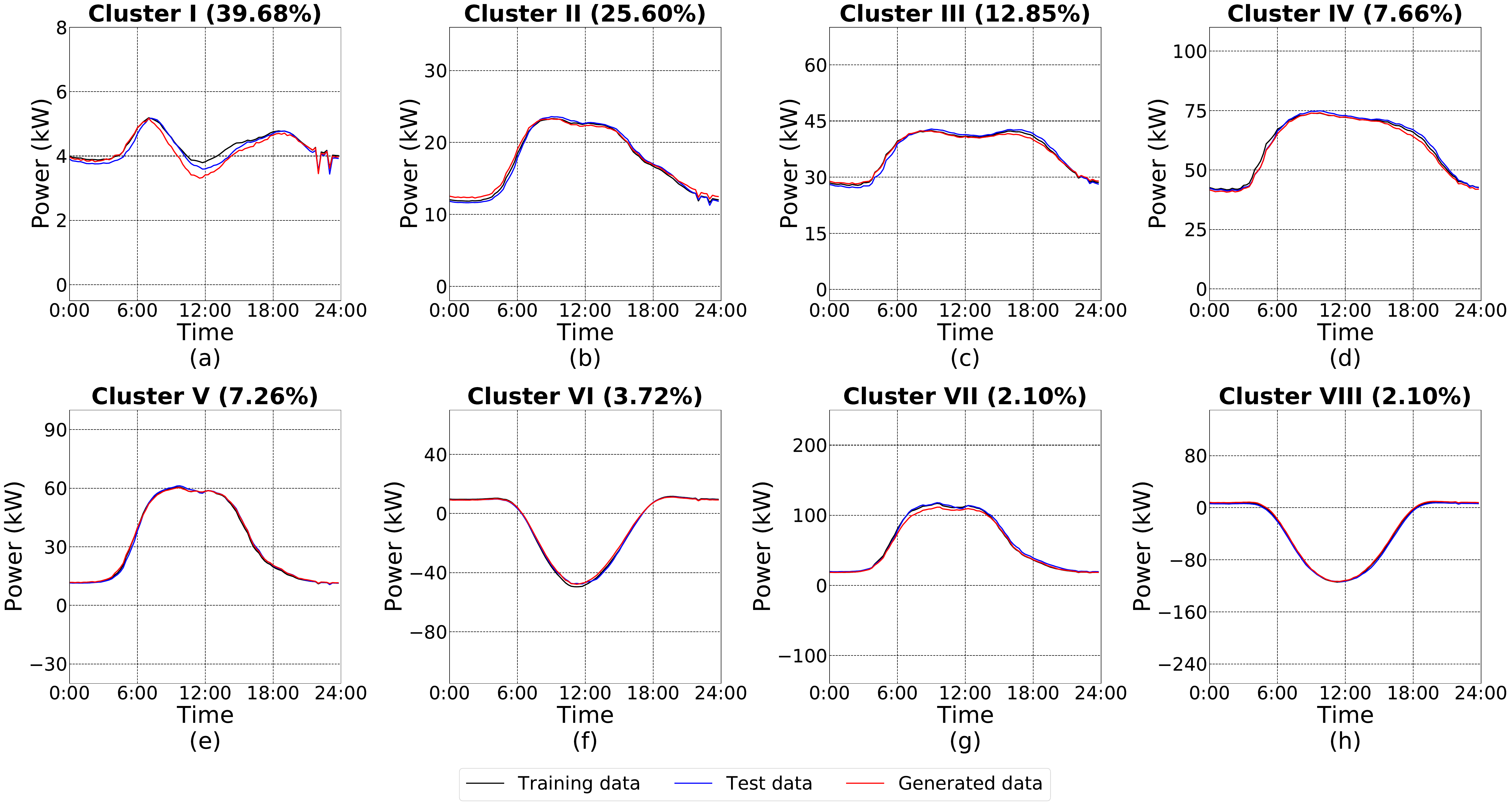}
	\caption{Mean value of historical and generated data in different clusters.}
	\label{fig:K-Means}  
\end{figure*}

The parameters of the generative models were tuned for optimal performance. The input and output layers had 96 dimensions (24 hours with 15-minute resolution). Accordingly, 96-dimensional daily load profiles were used for training and generation. The network contained 3 hidden layers in the encoder with dimensions of 800; the bottleneck layer had 12 nodes (12-dimensional latent vector). The decoder also had 3 hidden layers with dimensions of 800. The contextual condition $c$ consists of a 2-dimensional month condition and a 1-dimensional per-user power exchange intensity. 

The ReLU activation function was used, except for the generation of $\mu$ ($\mu'$) and $\sigma$ ($\sigma'$) leading up to the bottleneck and output layers. The \emph{adaptive moment estimation} (Adam) weight optimizer \cite{kingma2014adam} was utilized with default settings to iteratively optimize the value of weight matrices $W$ and bias vectors $b$. The batch size and learning rate parameter $\alpha$ for training were 1,280 and $10^{-5}$ respectively. The weighting factor $\beta$ was set as 8.5. Training and data generation of the model was conducted in Python using \texttt{tensorflow} on the Google Colab environment using the GPU option. The training process is shown in Fig.~\ref{fig:ISGT-training_tendency}. The Kullback-Leibler loss rapidly stabilizes during training. However, the reconstruction loss of the test data set starts to deviate from the training loss and fluctuates strongly after 1,000 training epochs, which indicates an overfitting of training data and general training instability. To find a compromise between loss minimization and generalization capacity of the trained model, 1,000 training epochs were used in this research. During the generation process, the total, monthly, and per user's amounts of the synthetic data are identical to the training set.

\section{Results}

\subsection{Comparison of daily load profiles}

To validate the generation capacity of our proposed CVAE model, we first visually inspect the generated contextual load profiles. We define the customers with the first and last 30\% of per-user intensities as small and large customers, respectively, and the remaining 40\% of users as medium users. In this experiment, we condition the generation of profiles on the months of April and July, and the `small' and `large' customer classes (random sampling of $c$ in their respective ranges). Fig.~\ref{fig:Load profiles} shows the mean value of load profiles under each condition combination (generated versus measured), and 10 randomly sampled load profiles alongside 10 random historical profiles. The mean generated load under each condition combination has a similar curve shape to the training data. Moreover, compared with historical data, the displayed load generations retain randomness and show a sense of realism, indicating that the CVAE model captures temporal features of historical load profiles.

\subsection{Clustering performance}
The following experiment compares all historical and generated daily load profiles for a more elaborate test of the distribution of generated load profiles. We first split the training data set into 8 clusters by the K-means algorithm \cite{lloyd1982least}, using the squared Euclidean distance metric. Then, we assign the generated and test load profiles to the nearest cluster. The mean values of training, test, and generated loads for each cluster are depicted in Fig.~\ref{fig:K-Means}a-h, in decreasing order of training data volume. The most voluminous cluster has small average load values. Note that the apparent gap in cluster I is smaller than the resolution of the data. Some clusters correspond to larger loads and generators (mainly solar PV). In all cases, the mean values of profiles assigned to the cluster match well.  

\begin{figure}[t!p]
    \centering
	\includegraphics[scale=0.225]{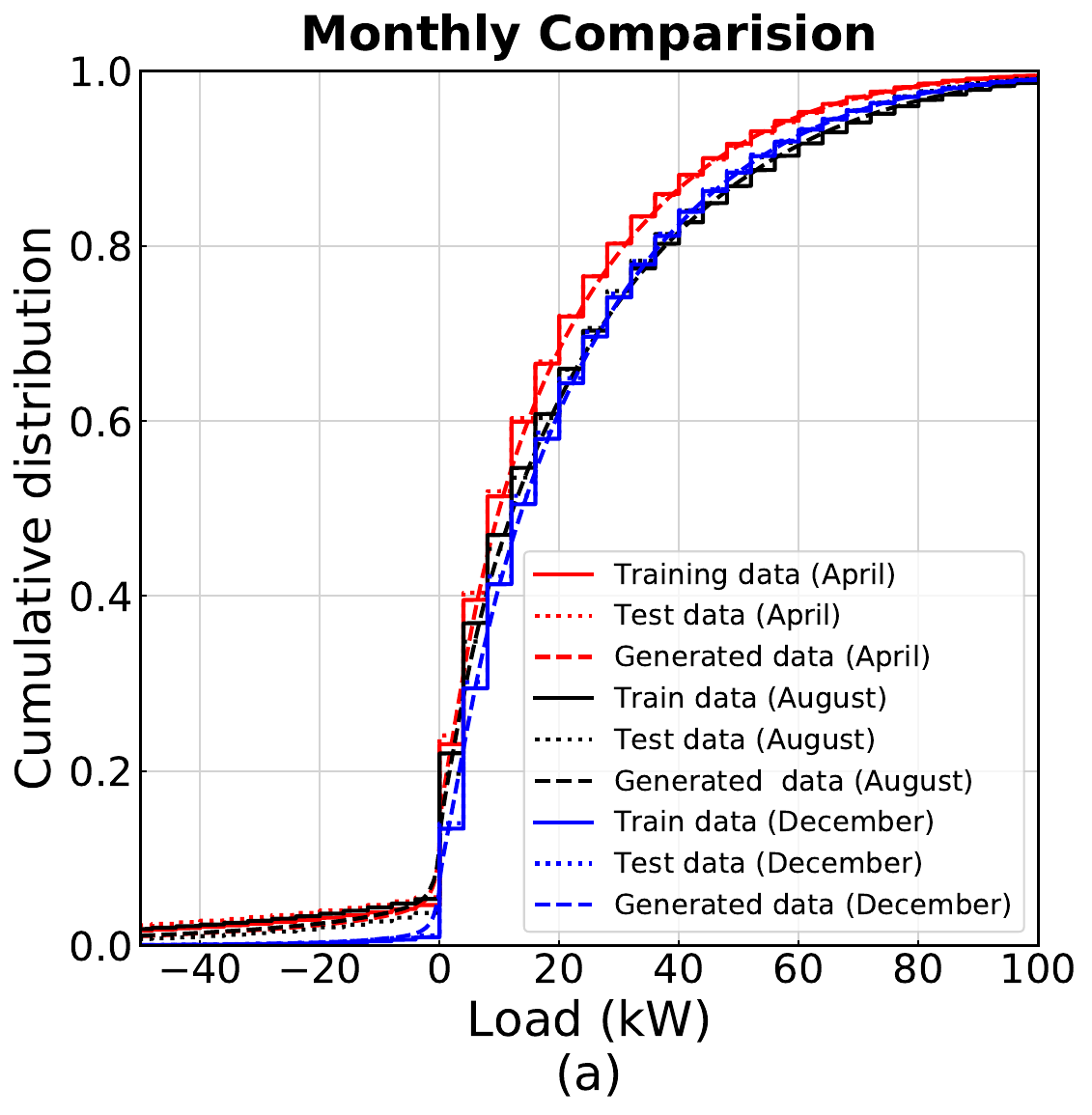}
	\includegraphics[scale=0.225]{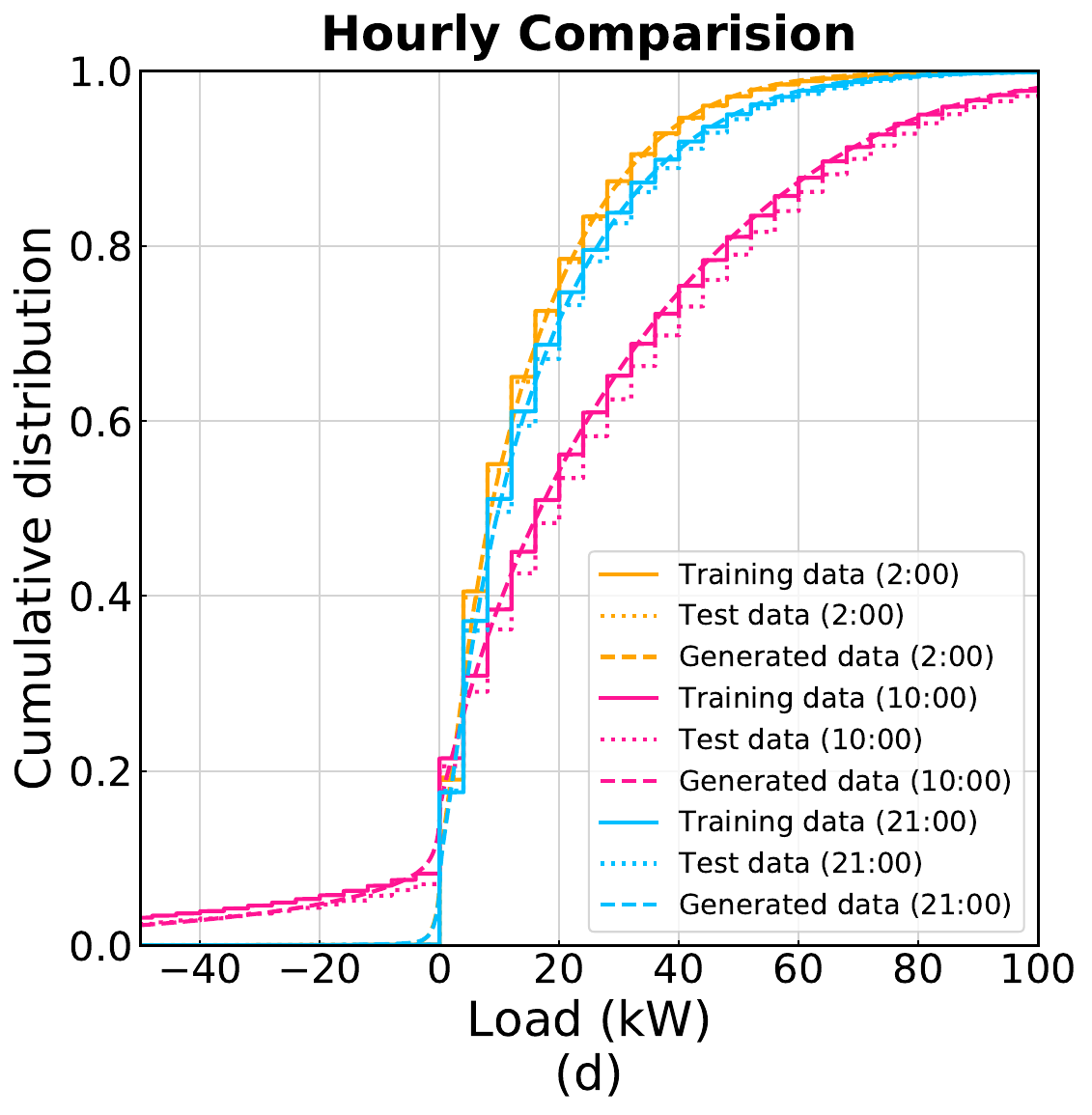}
	\includegraphics[scale=0.225]{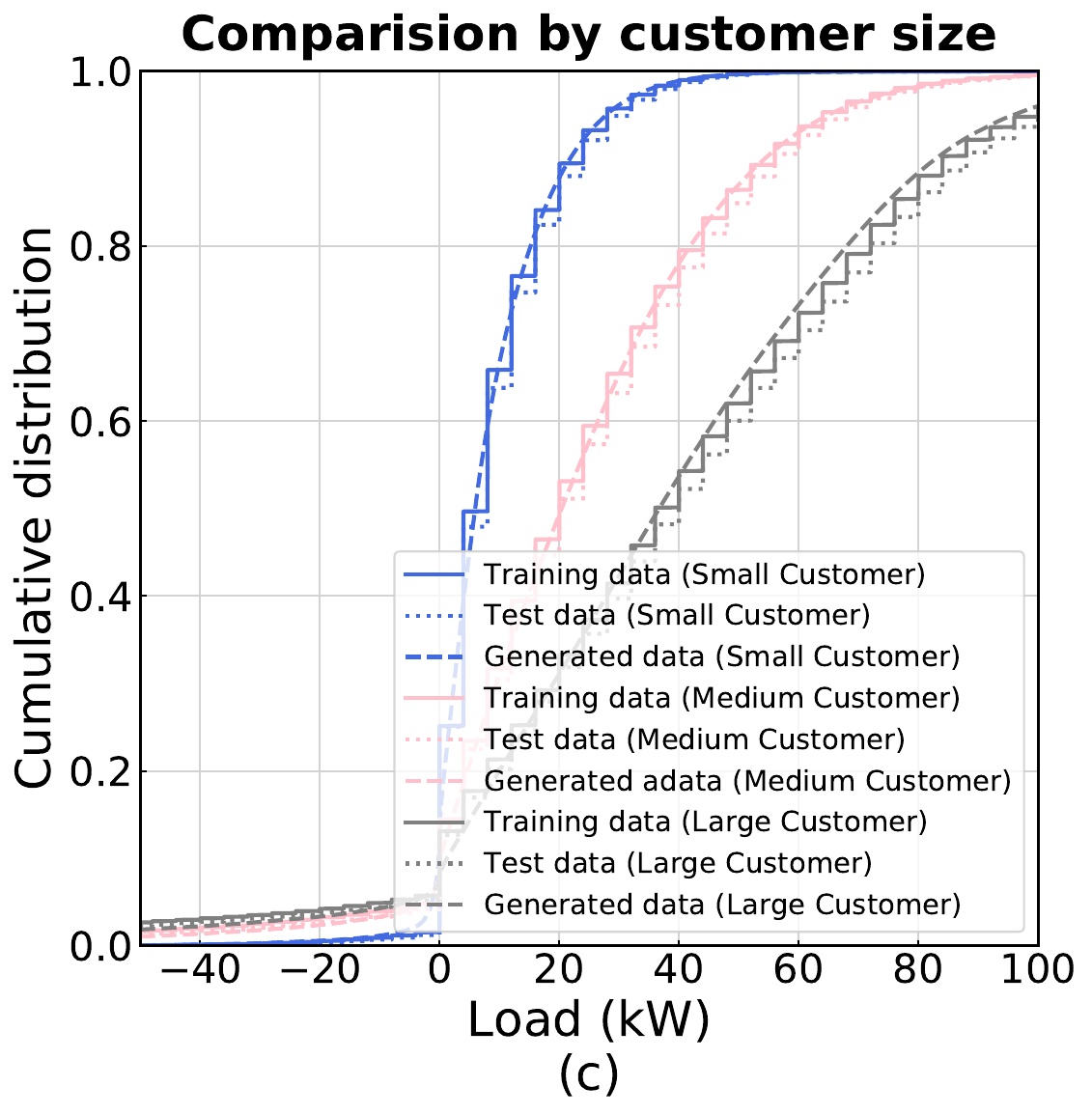}
	\includegraphics[scale=0.225]{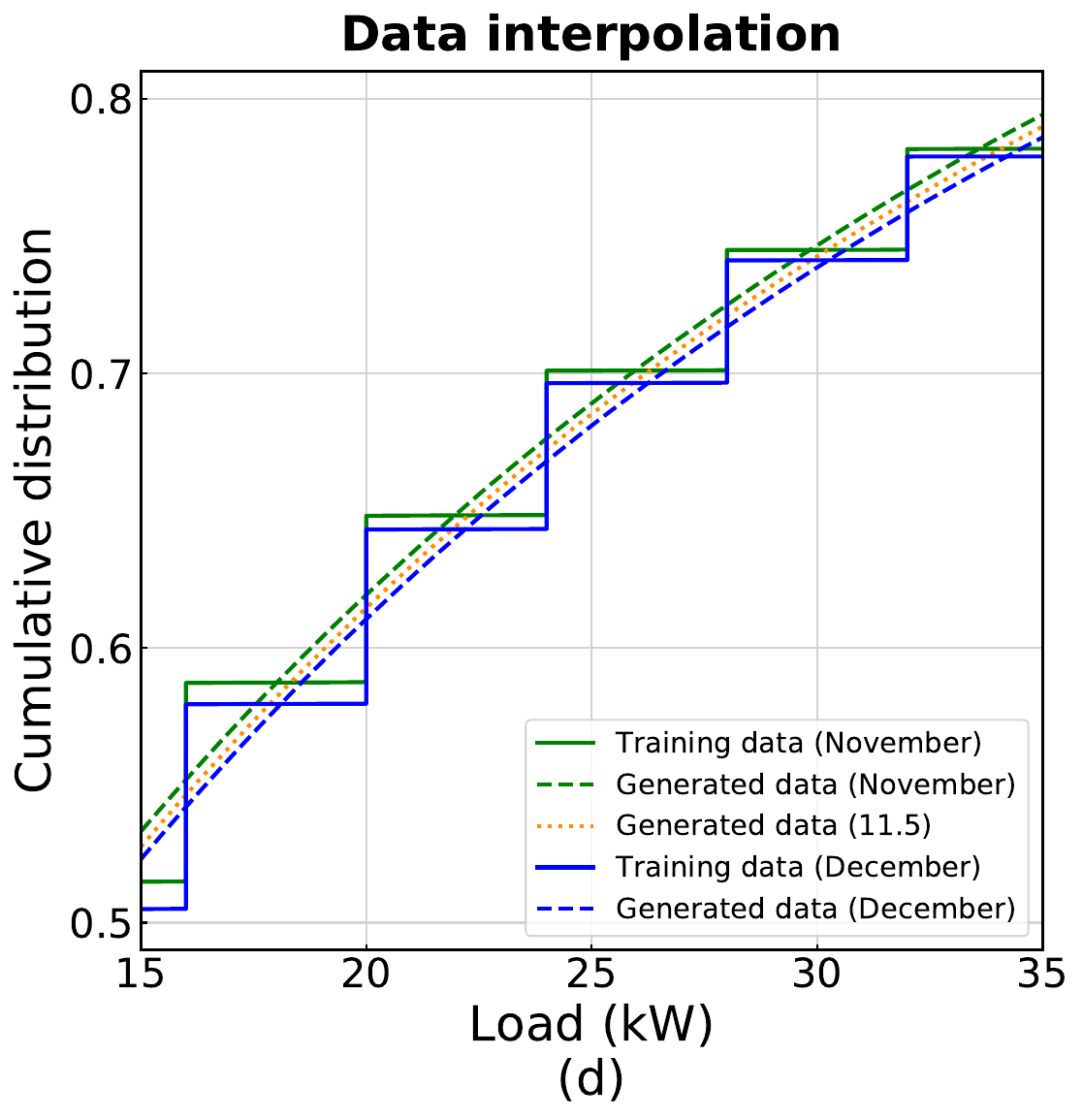}
	\caption{Cumulative distribution comparison of historical and generated data via different time scales and for users of various sizes. }
	\label{fig:CDF}  
\end{figure}

\begin{figure*}
  \centering 
    \includegraphics[scale=0.29]{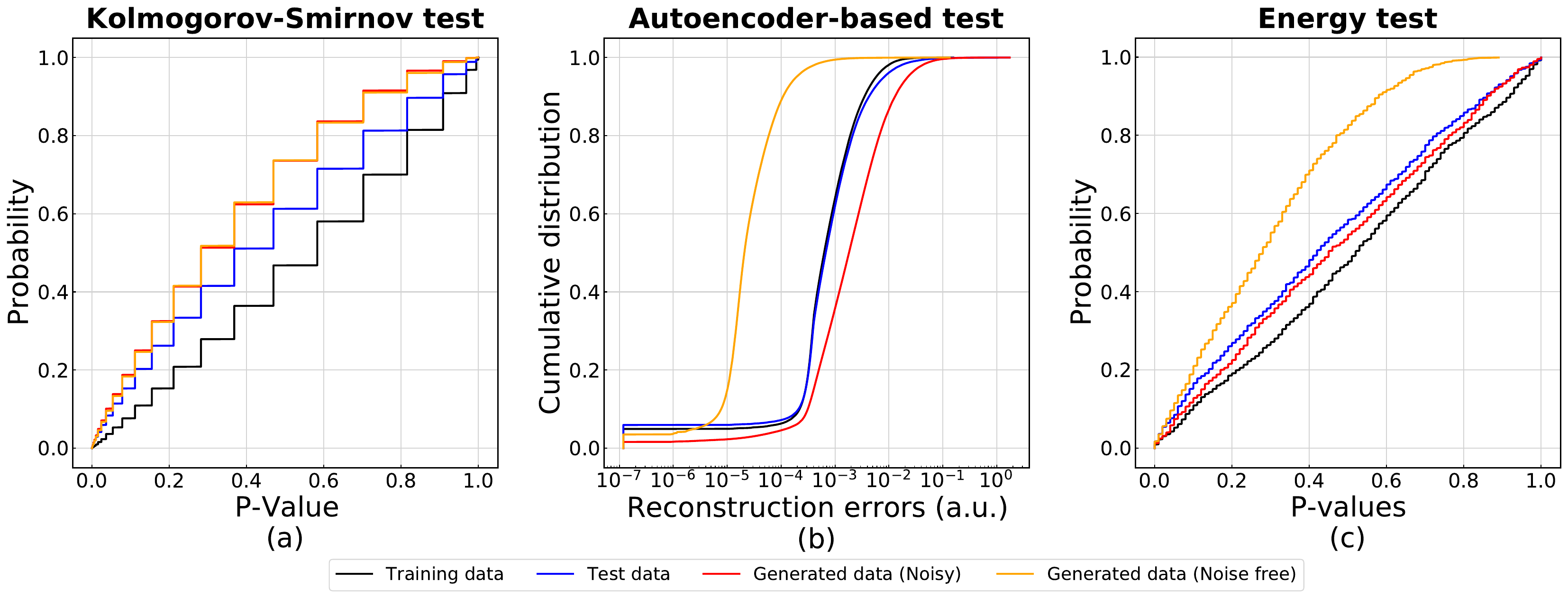}
  \caption{(a) Kolmogorov-Smirnov test. (b) Autoencoder-test. (c) Energy test}
  \label{fig:Statistical test} 
\end{figure*}

\subsection{Marginal distribution comparison}

The third experiment compares the cumulative distribution of historical and generated data via different time scales and users of various sizes. The experimental results are shown in Fig.~\ref{fig:CDF}; note the discretization of the real measurements, visible in these graphs. Fig.~\ref{fig:CDF}a exhibits the cumulative distribution of loads in different months. The CVAE model is able to generate contextual load profiles that follow the monthly distribution variation of historical loads. The hourly comparison of the load depicted in Fig.~\ref{fig:CDF}b shows that the curves of generations overlapped with the historical training data, demonstrating quite similar hourly distributions. Moreover, the comparison result shown in Fig.~\ref{fig:CDF}c stands for a good capture of load patterns of different customer sizes. Finally, we test the interpolation capacity of the CVAE model. Specifically, we use a virtual month condition (11.5) to generate load profiles, and the result is shown in Fig.~\ref{fig:CDF}d. The cumulative distribution of the load profiles with a month condition 11.5 lie between the distribution of loads in November and December. This demonstrates that the trained CVAE model can generate data using nonexistent conditions during the training process. Also, these profiles have features of data generated using nearby conditions.

\subsection{Statistical tests}

To further test the capacity of the CVAE model to generate realistic load profiles, non-visual statistical tests are implemented to inspect different aspects of generations. Specifically, in this experiment, the Kolmogorov-Smirnov test, autoencoder-based test, and energy test are utilized to examine univariate marginal distributions, point-wise multivariate dependencies, and multivariate dependencies of population, respectively. Interested readers can refer to \cite{wang2021generating} for more information on these tests. 
In addition to generations with noise $\epsilon \odot \sigma'(\Tilde{z},c)$ added (these were the data used in previous experiments), we also test the performance of commonly used noise free generations $\mu'(\Tilde{z},c)$ (see also the discussion in \cite{wang2021generating}).

Evaluating the performance on the Kolmogorov-Smirnov test (Fig.~\ref{fig:Statistical test}a), which assesses the accuracy of the marginal distributions, shows a small difference between the training and test sets, and a similar further difference in the distribution accuracy of the generated data. 
Comparing the results to those reported in 
\cite{wang2021generating} for country-level data, we see a slight degradation of the noisy generator. This could be because the individual load profiles are less smooth than the country-level snapshots, and a relatively large amount of synthetic noise $\epsilon \odot \sigma' (\Tilde{z},c)$ is added to base profiles $\mu'(\Tilde{z},c)$. This can result in the generation of extreme values, which reduces the test scores. 

The autoencoder test trains a separate (regular) autoencoder on the training data. This permits quantification of the quality of individual load profiles. The distributions of reconstruction errors obtained using real and generated data are shown in  Fig.~\ref{fig:Statistical test}b. The training and test patterns show similar distributions, and the `noisy' CVAE generates distributions with slightly worse reconstruction errors. However, the `noise-free' variation produces data that is significantly too smooth, resulting in reconstruction errors that are approximately two orders of magnitude lower. 

Finally, the energy test quantifies the similarities between high-dimensional \emph{distributions} of profiles. The results in Fig.~\ref{fig:Statistical test}c shows a similar performance between the generated profiles (noisy) and the test data, suggesting good generalization performance. Again, the generated data is a lot more realistic than when no noise is inserted in the output stage (`noise free').

\section{Conclusion and future work}

In this paper, we have investigated the capacity of the CVAE-based model to generate contextual load profiles, randomly selected across a mix of customers: pure loads, pure generators, and mixed load/generators. The load profile generator was trained on data from more than 4,000 industrial and commercial customers, and conditioned on the month of the year and the `size' of the customer, the latter being based on its power exchange with the grid on days with high grid usage. The experimental results demonstrate the model is able to generate visually realistic profiles and perform well on a number of statistical tests.
The results also reconfirm the importance of explicitly including (trained) noise in the final stage of the profile generator. 
In future work, we aim to refine our model to better control the production of extreme load values and more complex dependencies between subsequent moments in time.

\section*{Acknowledgment}

The authors are grateful to Alliander NV for providing anonymized load data of industrial and commercial users for this study.

\bibliographystyle{IEEEtran}
\bibliography{ISGT}

\end{document}